# Machine Learning and Manycore Systems Design: A Serendipitous Symbiosis

**Ryan Gary Kim**, Carnegie Mellon University
**Janardhan Rao Doppa**, Washington State University
**Partha Pratim Pande**, Washington State University
**Diana Marculescu**, Carnegie Mellon University
**Radu Marculescu**, Carnegie Mellon University

## Abstract
Tight collaboration between experts of machine learning and manycore system design is necessary to create a data-driven manycore design framework that integrates both learning and expert knowledge. Such a framework will be necessary to address the rising complexity of designing large-scale manycore systems and machine learning techniques.

## Introduction

Advanced computing systems have long been a fundamental driver in pioneering applications and technology, either through sheer computational power or form factor miniaturization. It has been no different in the emerging Big Data era. Large-scale datacenters have enabled complex machine learning algorithms to analyze and decipher massive amounts of raw data. Simultaneously, mainstream CPUs and GPUs have brought many of the lower complexity algorithms to the masses. These innovative models and learning techniques allow us to analyze and interpret large quantities of data, making it possible to exceed human decision making in multiple domains [1].

However, with the rising needs of advanced machine learning for large-scale data analysis and data-driven discovery, machine learning experts need low-cost, high-performance, and energy-efficient *commodity* manycore systems at their disposal. Developing these application-specific hardware must become easy, inexpensive, and as seamless as developing application software to keep up with the rapid evolution of machine learning algorithms. Therefore, it is of high priority to create an innovative design framework that reduces the engineering costs and design time of machine learning specific manycore systems. This framework will enable the *democratization* of access to application-specific hardware and make these systems widely available to machine learning researchers and data scientists.

To aid low-cost, energy-efficient, and small form factor implementations for the current applications of interest (including machine learning), computer architects have strived to increase the level of parallelism on manycore systems. This increased level of on-chip parallelism will further improve our ability to run machine learning algorithms and Big Data analytics commensurate with the number of cores on the chip. However, these highly integrated manycore architectures cause additional problems. As the number of cores increases, system complexity and the number of interdependent design decisions grow, thus escalating the need for a *holistically optimized* design process that makes design decisions across all layers of the system (subsystems), *e.g.*, memory, compute, and network. Additionally, rising levels of variability within the manufacturing process and system workload has made it increasingly difficult for manycore systems to quickly adapt to changing system requirements. This volatility necessitates a movement towards *fully-adaptive* systems [2], where all subsystems are collectively tuned to optimize the overall efficiency. Both the increasing system design complexity and operational variability have made it increasingly difficult to explore the expanding combinatorial design space and optimize manycore systems.

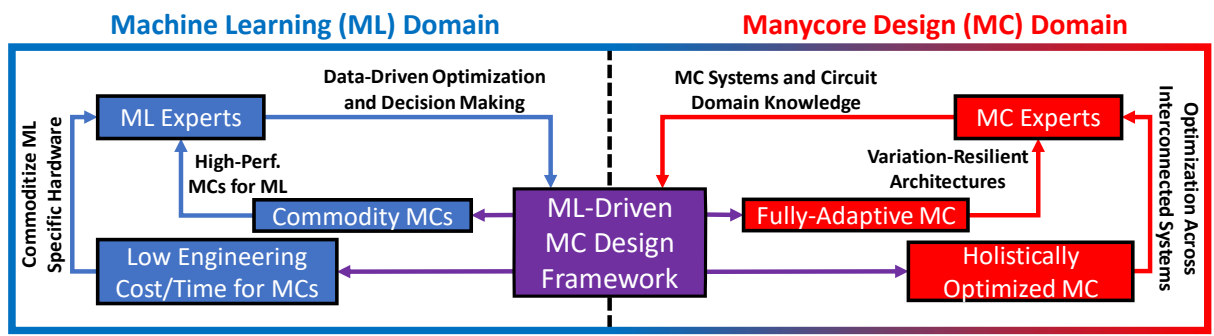

Fig. 1. Tight collaboration between Machine Learning (ML) and Manycore (MC) specialists enable a valuable interdisciplinary effort to create manycore systems that will empower the next wave of machine learning algorithms and manycore design methodologies.

To address these challenges, we believe machine learning can provide a natural solution to create application-specific, energy-efficient computing systems with significantly less engineering effort. Indeed, machine learning techniques have been increasingly utilized to intelligently explore the system tradeoffs (*e.g.*, performance, energy, and reliability) within the design space to quickly uncover near-optimal candidate manycore designs [3]. In short, machine learning approaches enable data-driven manycore system optimization and provide the necessary computing and planning support for design space exploration of the system architecture and online control policies.

Through tight collaborative efforts between manycore system designers and machine learning experts, we can create a framework that espouses manycore system design knowledge and data-driven decision making. This interdisciplinary effort will greatly benefit both machine learning and manycore experts. Some examples of this interdependency between machine learning and manycore design are shown in Fig. 1. This framework will allow us to create *fully-adaptive* systems that are *holistically optimized* across the entire design stack. Manycore system designers will gain insights by understanding the rationale behind the machine learning-driven manycore design decisions, an example of "data-to-insight" for manycore design. By significantly reducing the engineering effort and cost of designing manycore systems, this design framework will enable the *commoditization* and *democratization* of massive application-specific manycore systems. Thus, making high-performance, application-specific manycores readily available to the rapidly changing machine learning domain.

Looking ahead, the amount of data at our fingertips will continue to explode and necessitate this mutually beneficial collaborative relationship: machine learning and manycore systems will need to inspire and motivate each other to continue innovation in their respective domains. This close collaboration can stimulate and empower the next wave of machine learning algorithms and manycore design methodologies.

## 1) MANYCORE SYSTEMS DESIGN CHALLENGES

The manycore (r)evolution continues to unfold, so hundreds of cores within a single chip will soon become pervasive. Such manycore chips will provide a level of parallelism traditionally reserved for large computing clusters. However, to make large single-chip manycore systems a reality, several fundamental challenges need to be addressed: 1) Exploring the huge design space and system tradeoffs among highly interdependent subsystems; 2) Satisfying thermal constraints that prevent parts of the manycore systems from being powered-on simultaneously, *i.e.*, dark silicon; and 3) Adapting to fluctuations in system parameters that occur at the transistor, subsystem, and application levels of the system. Before we can understand how machine learning techniques can enable the design of holistically optimized and fully-adaptive manycore systems, we must briefly discuss the problem space of manycore system design.

### 1.1) Design-Time Optimization

At design-time, the various optimization decisions for manycore systems can largely be separated into several layers of the platform:

1. **Compute Layer** – The compute layer contains the compute cores of the system. Design parameters include: Compute core architectures (CPU, GPU, FPGA, ASIC); Instruction set architectures (RISC, CISC, VLIW); Number and blend of compute cores (homogeneous – single core type, heterogeneous – multiple core types).
2. **Interconnection Layer** – The interconnection layer maintains the communication infrastructure to enable coordination between the compute cores in the system. Design parameters include: Topology (*e.g.*, bus, ring, mesh, irregular, 3D); Link type (*e.g.*, wireline, wireless, optical, Through-Silicon Via (TSV)); Router design (*e.g.*, number of stages, virtual channels, arbitration mechanisms).

3. **Memory Layer** – The memory layer is the subsystem that coordinates data accesses in the system. Design parameters include: Hierarchy (structure from compute cores to main memory); Distributed or shared memory; cache coherency protocols; Emerging memory technologies.

Existing design methodologies typically optimize the above architectural parameters to serve a particular design scenario, *e.g.*, high-performance, low-power. In addition, many of these parameters provide tuning knobs to fine-tune the tradeoffs between various objectives. However, in order to achieve a *holistically optimized* manycore system, all of these parameters and tuning knobs must be considered and optimized together.

### 1.2) Run-Time Control Optimization

After the initial design of the system, the platform must be able to adapt to conditions seen during run-time, *e.g.*, application characteristics, process-variations (chip-to-chip variability), and aging components. The problem space for run-time optimization can be largely categorized into:

1. **Compute Management** – Dynamic control of the compute layer parameters. Various compute management problems include: Dynamic voltage and frequency selection; Process-variation-aware available frequency range; Reconfigurable core configuration.
2. **Interconnection Management** – Dynamic control of the interconnection layer parameters. Various interconnection management problems include: Dynamic voltage and frequency selection; Adaptive and prioritized routing; Medium access control.
3. **Application Management** – Application adaptation to current system characteristics. Various application management problems include: Task mapping (Task to core mapping); Task scheduling; Kernel selection (algorithm and level of parallelism); Precision/approximability (application sensitivity to output precision).

Many control techniques, including some based on machine learning [4][5], have been developed to handle the static (*e.g.*, manufacturing generated) and dynamic (workload-driven, aging, thermal) variations. However, many of these solutions are ad-hoc, not scalable, or require significant human decision making. To move towards *fully-adaptive* manycore systems, much of these decisions should be replaced with automated data-driven decision making implemented through machine learning.

### 1.3) Optimization Objectives

Given this large space of optimization and tuning parameters, engineers and designers are then given the daunting task of creating a system that is able to satisfy both system constraints and functionality requirements. Additionally, given the variability of the system requirements due to various scenarios (*e.g.*, application characteristics and mix of applications, input data set, process variations) while running the manycore system, each system layer and run-time manager needs to be carefully chosen. Some of the most popular design scenarios include:

1. **High-Performance Computing (HPC)** – Designs for

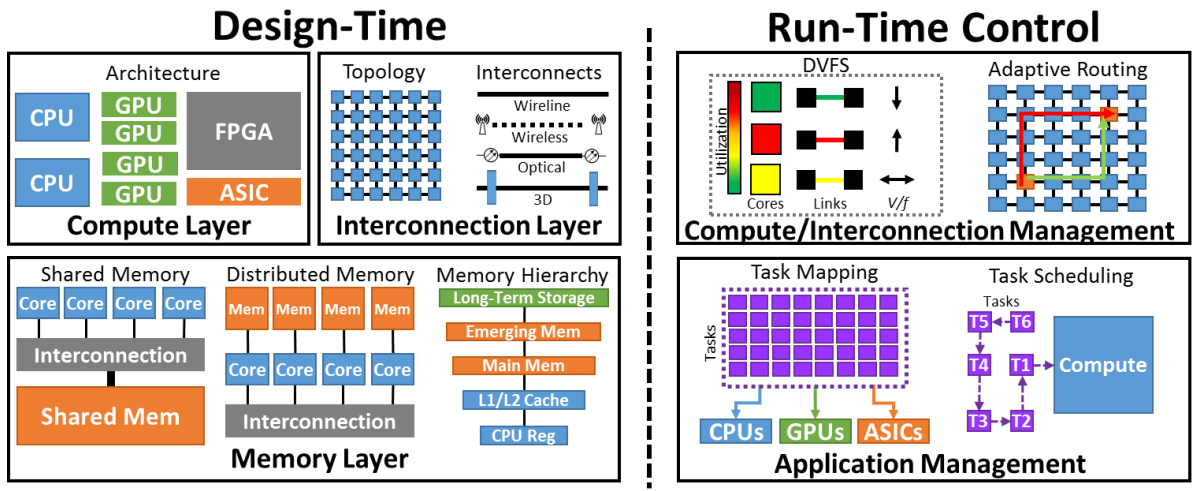

Fig. 2. A small sample of design parameters in consideration for each system. Several examples are given for both design-time and run-time decisions at various sections of the system. Each system must be carefully chosen depending on the application scenarios.

HPC need massive levels of parallelism (*e.g.*, manycore systems, supercomputers) to handle large, computationally demanding tasks. These system designs are mostly concerned with overall performance, performance per watt, thermal profiles, and implementation space (*e.g.*, sq. ft. taken in the computing center).

2. **Mobile/Embedded Computing** – Designs for mobile computing are expected to handle the workloads typically seen on mobile devices, while embedded computing is expected to efficiently handle dedicated functions. These system designs are mostly concerned with reliability, meeting real-time constraints, battery life, and thermal profiles.
3. **Internet of Things (IoT)** – Designs for components of the IoT are expected to coordinate the sensors, wireless communication, and actuators that make up the core of most IoT computing paradigms. These system designs are mostly concerned with battery life, idle power draw, form factor, and system lifetime.

Clearly, the design space (combination of design-time decisions and run-time policies for a set of target application and optimization scenarios) is exploding as we progress through the manycore era. Fig. 2 shows a small sample of design parameters available today, *e.g.*, core architecture, memory architecture, task mapping, adaptive routing. Current engineering techniques tend to reduce the dimensionality of this design space by considering only a few of these optimization decisions in isolation. For example, optimizing only the compute layer or the network+memory layers [6]. These isolated optimization techniques result in sub-optimal designs and prevents us from achieving holistically optimized systems.

This high dimensional design space will only grow in size and complexity as new technologies emerge. Each system must be carefully chosen from this considerably large design space depending on the application scenarios envisioned for the platform. The difficulties of engineering within this design space, alongside greater process and application variability, will prevent efficient human decision making for optimizing the manycore design.

### 2) ADVANCED MANYCORES AS AN ENABLER FOR MACHINE LEARNING APPLICATIONS

As computing capabilities continue to improve and these manycore systems become more ubiquitous, more complex algorithms and applications will become feasible. For instance, state-of-the-art deep learning algorithms that take days to complete would have taken years to train on past platforms. Likewise, in 2013, [7] demonstrated that clusters of servers containing manycore GPU chips could train large deep neural networks in a few days with significantly less resources. By creating an advanced manycore design framework, we will allow machine learning researchers to assess a wide range of cutting-edge heterogeneous manycore systems that adapt to algorithmic changes. This framework will establish a direct link between machine learning and manycore systems that will form a positive feedback loop that accelerates improvements in computing and algorithms.

In what follows, we highlight three important machine learning applications that enable data-driven decision-making where enhanced manycore systems will have a significant impact: 1) Deep learning techniques that allow us to learn appropriate representations from raw data; 2) Learning models from large-scale data with appropriate representation; and 3) Rational decision-making to control complex systems. Fig. 3 shows some of the key machine learning capabilities (in red) that will be extended due to larger, more efficient manycore designs. Increases in the depth of deep neural networks, the volume and diversity of data and their sources, and the size of lookahead search trees for automated planning heavily rely on enhancements in computational power. This motivates the need to create low-cost, high-performance *commodity* manycore systems specialized for machine learning algorithms. Simultaneously promoting the *democratization* of access to such systems to machine learning experts.

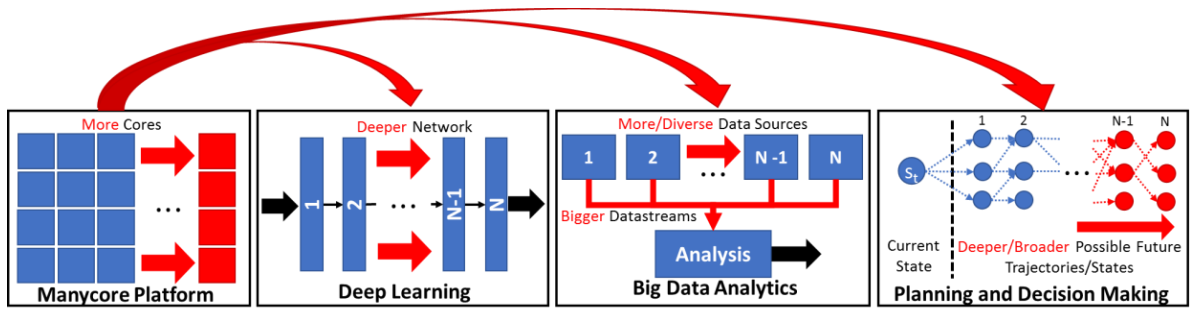

Fig. 3. Increasing manycore size and capabilities allow further innovation in machine learning techniques (in red). For example, larger manycore systems can enable larger and more complex deep learning network architectures, data analytics on larger and heterogeneous sources of data, and planning+decision making based on deeper and broader lookahead search trees.

### 2.1) Deep Learning

In recent years, deep learning techniques have seen great success in diverse application domains including speech processing, computer vision, and natural language processing. While the fundamental ideas of deep learning have been around since the mid-1980s, two key reasons for their recent successes are: 1) Availability of large-scale training data; and 2) Advances in computer architecture to efficiently train large-scale neural networks using this training data.

Simply speaking, deep learning refers to a class of machine learning algorithms, where the goal is to train a layered neural network using training data. By exploiting the parallelism existing between computations in the same layer, deep learning operations can be significantly accelerated through manycore architectures. Recent work has taken advantage of this parallelism and unique computational requirements to design custom cores to accelerate deep neural networks (NN) [8]. Google has even found the need to design custom ASICs [9] to accelerate the inference portion of the few NNs that make up a majority of their current datacenter NN workload.

In the future, the scale of data and the size of the neural networks will grow even further. Additionally, due to algorithmic advances, the computational nature of these algorithms may change. Therefore, a framework to aid new architectural and optimization innovations will be needed to design high-performance and energy-efficient manycore systems to meet the growing requirements.

### 2.2) Large-Scale Machine Learning via MapReduce

The data processing associated with many machine learning algorithms can be mapped into MapReduce tasks to leverage distributed computing resources to perform training over large-scale datasets [10]. Recent work has demonstrated that using a wireless Network-on-Chip (NoC) enabled VFI-based manycore system, it is possible to improve the energy-efficiency of MapReduce implementations without paying significant execution time penalties [11]. However, in this case, the specific subsystems were hand-tailored to the specific MapReduce implementation. It is our vision to have an automated framework that will consider all subsystems and reduce the amount of manual engineering effort.

### 2.3) Planning and Large-Scale Decision Making

Automated planning and decision making under uncertain circumstances has many real-world applications, including logistics, emergency response, resilient power grids, and ecosystem management for sustainability. We are increasingly witnessing domains that are modeled as large-scale (Partially Observable) Markov Decision Processes, where we need to make high-quality decisions under tight time constraints [12].

Monte-Carlo planning has emerged as a very promising technique to solve these kinds of problems. Upper confidence tree (UCT) is a popular online planning algorithm which selects an action at a given state by building a sparse lookahead search tree over the state space with the given state as the root [13]. UCT incrementally constructs the tree and updates action values by carrying out a sequence of Monte Carlo rollouts of entire decision-making sequences from the root to a terminal state. The key idea behind UCT is to intelligently bias the rollout trajectories toward those that appear more promising based on past experience while maintaining sufficient exploration. In [14], the authors investigated the scalability of Monte Carlo Tree Search on Intel Xeon and Intel Xeon Phi processors. Significantly, they not only noted the importance of the number of cores, but also the processor architecture in the performance of the manycore system. By designing a manycore system for Monte-Carlo planning, we can maximize the number of rollouts per second and produce higher quality decisions in less time.

### 3) MACHINE LEARNING-DRIVEN MANYCORE DESIGN FRAMEWORK

Due to the wide-spread availability of computing power and data, machine learning has recently seen a boom in popularity, solving complex problems in diverse application domains. As the "Evolution to Fully-Adaptive Designs" sidebar advocates, machine learning is also well-suited to govern the data analysis and decision making within and between the subsystems to realize the vision of fully-adaptive systems and holistically optimized manycore architectures. Additionally, we not only advocate machine learning as an enabling framework to design these manycore systems, but that these advancements will reduce the engineering time and cost in designing manycore systems while providing the needed computing capabilities to advance machine learning algorithms. This framework will

> **EVOLUTION TO FULLY-ADAPTIVE DESIGNS**
>
> Until recently, engineers have followed a deterministic approach by designing systems assuming a worst-case scenario. For years, many aspects of microarchitecture design (*e.g.*, performance, power, and thermal trade-offs) were governed by this worst-case paradigm. However, with the continuation of transistor scaling into the ultra-deep sub-micron, assuming deterministic constraints for design parameters became inadequate [1] and design paradigms had to adapt. For example, design methodologies had to consider noise and variability within the input parameters and create variation-tolerant designs. Thus, computer system design began to transition from a deterministic to a probabilistic design paradigm. With the advent of manycore chips, the design of the uncore (functionality outside the processing cores) subsystems became more complex. Operations associated with the "uncore" [2] had to coordinate their efforts to ensure reliable operation of the whole system. Designers had to consider many probabilistic measures in the face of uncertain application and technology variations.
>
> Unfortunately, due to the static nature of designs with deterministic and probabilistic design parameters, many opportunities to maximize design goals are missed. To take advantage of these opportunities, designs have evolved to adapt to changes from within and outside the system, and allocate precisely the necessary resources. However, this paradigm of adaptive computing is in its infancy relative to the development of the rest of the processor, with most designs optimizing one or two objectives among a few subsystems. Instead, moving forward, the adaptive computing paradigm needs to take a holistic approach to achieve maximum efficiency. By learning the ever-changing internal (power, performance, thermal, wearout, etc.) and external (application data, workload mix, etc.) characteristics of the system, accompanied with intelligent system adaption using this information, significant improvements can be made to maximize the design's "utility."
>
> However, as these systems continue to grow increasingly complex, the number of design and control decisions for manycore chips become intractable. The number of sensors and actuators that will be needed for adaptive computing will not only grow, but operate in an increasingly complex coordinated effort to maintain optimal utility. Making sense of all this input data and design decisions becomes a fundamental problem in computer architecture design. Here, Machine Learning presents an especially attractive solution to solve this problem in manycore systems in a data-driven fashion.
>
> [1] S. Borkar. "Microarchitecture and design challenges for gigascale integration," *in Proceedings of the International Symposium on Microarchitecture*, Portland, OR, USA, 2004.
> [2] S. Borkar, "Thousand core chips: a technology perspective," *in Proceedings of the 44<sup>th</sup> Annual Design Automation Conference*, San Diego, CA, USA, 2007.

pave the way to commoditizing and democratizing single-chip manycore systems to spur machine learning research among the masses.

Ideally, this framework would be able to inform the manycore designer of the best system configuration for a particular set of applications and operating configurations. For example, a manycore GPU system could be the best high-performance system for MapReduce. However, this optimal point may change under differing power or area constraints. By providing this optimization in an automated manner, we can quickly configure and design manycore systems for a wide-range of application and operating scenarios.

Through tight collaboration, this design framework will combine the benefits of data-driven decision making enabled by machine learning techniques with the vast amount of experience and prior knowledge from manycore researchers (to serve as inductive bias [15]). This will enable the design and optimization of manycore systems at different levels of abstraction, broadly classified into static and dynamic optimization. In what follows, we will discuss these optimization problems along with examples, and explain how machine learning techniques with manycore expert guidance can be employed in solving them effectively.

### 3.1) Static Optimization

Optimization processes that are carried out only once in the design process qualify as *static optimization* problems. One important example is the design-time optimization of manycore systems for real-world applications, such as the design of Intel processors using their "tick-tock" model. Each "tick" refers to the shrinking and optimization of their process technology and each "tock" refers to a new architecture design. By using the knowledge gained on older technology, they transfer the architecture to a new technology; then they incrementally change the architecture by reusing as much as possible. Each of these ticks and tocks are normally separated by a year, creating a design paradigm that resembles an iterative and greedy search process that relies on the guidance from human designers. By automatically learning from available data, machine learning can guide the manycore experts toward much more innovative design decisions to speed up the manycore systems progress.

Machine-learning techniques can be useful in learning knowledge from past problem-solving experience to find high-quality solutions faster than non-learning based algorithms. In short, machine-learning enables the problem-solver (a computational search procedure) to make intelligent search decisions to achieve computational efficiency for finding (near-) optimal solutions.

**Optimizing Known Cost Functions:** In many cases where the optimization cost function is known (*e.g.*, network connectivity, place-and-route), the challenge lies in finding high-quality solutions within prohibitively large design spaces. Simulated annealing (SA) is a heuristic search algorithm that has been able to approximate the global optimum in a large space. The amount of time needed for SA to perform well will quickly become prohibitive as our design space continues to expand. Fortunately, an online machine learning algorithm, STAGE, allows designers to intelligently explore the design space [16]. For example, recent work adapted STAGE to optimize the NoC for manycore systems and showed significant speedup results over SA and genetic algorithms [3].

STAGE was originally developed to improve the performance of local search algorithms (*e.g.*, hill climbing) for combinatorial optimization problems. Fig. 4a shows a high-level representation of STAGE. STAGE attempts to learn a model that maps features of a starting state of the search procedure to the starting state's potential of finding a high-quality solution. The effectiveness of the learned model depends on a small subset of critical training examples that successfully teach how to avoid different local optima. Manycore researchers can begin by guiding or biasing the model by indicating what features are important in the manycore design space, greatly increasing the efficiency of STAGE.

Optimizing Unknown Cost Functions: In some cases, we may not know the cost functions to perform the optimization. To evaluate the quality of a candidate design, we need to perform computationally-expensive simulations. For example, optimizing the design for chip lifetime: this is a very challenging scenario that cannot be handled without the use of machine learning techniques.

Bayesian Optimization (BO) is applicable for this setting [17]. BO algorithms are sequential decision-making processes that select the next candidate design to be evaluated to quickly direct the search towards (near) optimal solutions by trading-off exploration and exploitation at each search step. By iteratively selecting a candidate design and learning a statistical model based on the observed input design and output quality pairs, the BO framework can quickly move towards an optimized design; this significantly reduces the amount of computationally-expensive simulation calls made during the design process. Similar to the previous example, manycore experts can also integrate their prior knowledge to bias the candidate selection process to known favorable solutions.

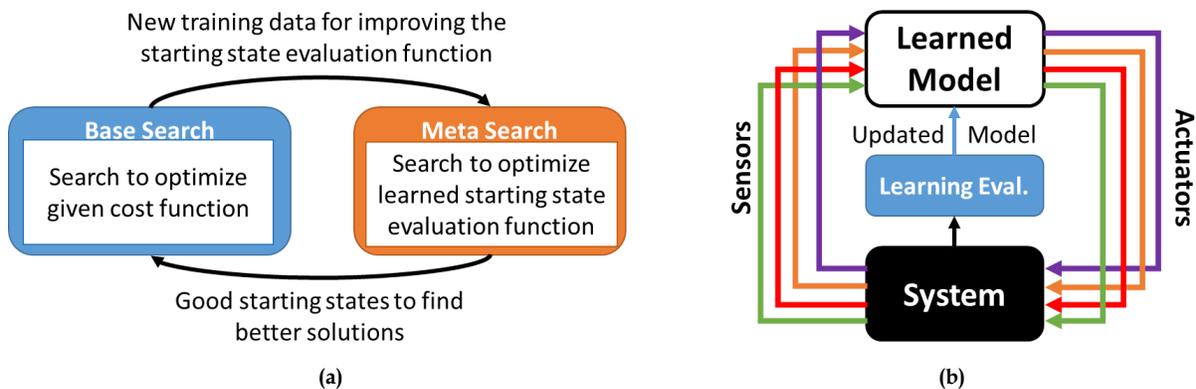

Fig. 4. Examples of how machine learning helps make data-driven decisions for manycore system design and control. (a) In classical search algorithms for manycore systems design, the starting state has a heavy influence on the quality of the output state. By learning the characteristics of good candidate starting states through a learned starting state evaluation function, machine learning can significantly speed up these search algorithms. (b) In the dynamic setting, machine learning can be utilized to learn a model that takes sensor outputs as input and output actuation signals that optimize an objective. This can be a continuous process to fine-tune the learned model during run-time to optimize the system.

### 3.2) Dynamic Optimization

Optimization processes that are carried out many times during the operation of manycore systems qualify as *dynamic optimization* problems. Two major objectives are typically considered when designing dynamic manycore policies: Creating a policy that can adapt to changing circumstances and creating a way for the policy to accurately predict how each action will affect the system. Even if the policy is able to "perfectly" act, if the online predictors for the action's effects are inaccurate, the results will be suboptimal.

In these cases, machine-learning techniques can be useful in learning the models and policies for dynamic manycore optimization by mapping the information from sensors at various levels of the manycore architecture to actionable decisions. This is especially useful when incorporating information that may not be straightforward and intuitive or relies on the decisions of multiple actuators and have intricate dependencies. Additionally, the policy can dynamically learn how its actions affect the system and readjust its model accordingly. We discuss a couple of examples related to dynamic optimization within manycore systems and how machine learning has been used to solve these optimization problems.

*Dynamic Power Management:* To maximize energy efficiency while meeting user-specified demands or constraints, modern processors and manycore systems must be equipped with a wide range of voltage and frequency values. However, the performance/power tradeoffs are heavily dependent on process variations, application characteristics, voltage, and frequency values, among other system aspects. By learning a model that accurately characterizes the power consumption as a function of the operating frequency under different workload behaviors and process variations [18], only then is it possible to accurately maximize the performance by manipulating the voltage and frequency of the compute layer within a power budget.

To combat scalability issues in power management, Voltage Frequency Island (VFI) group cores and communication links together and use a single power manager to significantly reduce the area overhead in the control module and power delivery circuitry. However, this significantly complicates the voltage and frequency tuning, as a single decision must be made for the entire group of cores and communication links. Machine learning can be utilized here to help generalize a model that isn't obvious, to efficiently allocate the voltage and frequency of a VFI [19].

*Adaptive Routing:* Within the on-chip interconnection system, it is important to monitor the levels of utilization across the network in order to make preemptive and corrective decisions to ensure high throughput and minimize congestion within the network. Using this information, we can consider adaptively rerouting the data to ensure a high functioning interconnection system. By dynamically learning which routing decisions effectively load-balances the network, we can create an efficient routing mechanism. Additionally, in emerging three-dimensional integrated circuits, an important parameter is the wear-out of through-silicon vias (TSV) that vertically connect the different dies. Similar to normal adaptive routing, we can dynamically learn a routing mechanism that load-balances the TSVs and maximizes the chip lifetime [20].

### 3.3) Towards Holistically Optimized, Fully-Adaptive Manycore Systems

The examples presented above and most other current solutions consider only a part of the system with one or two optimization objectives. This is perfectly fine as a point solution and such works can be leveraged when considering full-system optimization. Unfortunately, when multiple contending objectives are considered, any optimization not considering the full-system will have incomplete information of the inter-layer interactions and tradeoffs associated with the design objectives.

To complicate things further, the design objectives may have varying degrees of importance and constraints, thus, the tuning knobs within all layers of the system need to be properly configured to achieve an optimal system. Without considering the entire system, one cannot hope to achieve the proper optimal tradeoffs available. Similarly, the authors in [2] advocate the high-level idea of cross-layer sensing and actuation to achieve a similar goal.

Fortunately, data-driven machine learning presents a way to learn how to select methodologies and tradeoffs for both design-time and run-time optimization of the system. Ideally, manycore system design should rely on using both design-time and run-time decisions to holistically optimize the system. In the short-term, existing regression analysis or neural networks can be used to find relationships between the design choices and the numerous optimization objectives. However, in the long-term, given this large and increasing state space, new techniques should be developed for design space exploration guided by hardware simulations and should be able to incorporate domain knowledge. For example, we can include qualitative statements that describe monotonic influences from the input to the output [21], such as "higher frequencies are more likely to produce lower execution time." This information should be utilized to restrict the state space search to practical and desirable solutions.

The run-time optimization should become fully-adaptive (Fig. 4b) where the system is monitored by many sensors that feeds into a learned model that actuates the control modules within the system. Periodically, the learned model can be evaluated and adjusted to adapt to current operating conditions. In the short term, deep neural networks, reinforcement or imitation learning can be applied in an online setting to learn these control policies with enough training [2][4][5][19][22]. However, in the longer term, more advanced techniques that involve modeling and abstracting the application characteristics, and using that information to significantly improve the control policies should be investigated.

## 4) CONCLUSION

In recent years, there has been much focus on the acceleration of machine learning and deep learning algorithms through manycore optimization. On the other hand, we have only scratched the surface of manycore optimization using machine learning techniques. Indeed, in current solutions, machine learning has shown promise in identifying high-quality candidate manycore designs. However, most modern manycore design techniques that utilize machine learning only support the optimization of a few objectives within a part of the system. In this sense, we are only at the beginning of this paradigm shift towards data-driven manycore system optimization.

Additionally, it has not been sufficiently explored how manycore designers can use their significant domain knowledge to guide the data-driven nature in machine learning to sensibly explore the design space for both architecture and run-time control, especially in the presence of variability caused by applications, emerging architectures and bleeding-edge process technology. By integrating this manycore expert knowledge with data-driven machine learning, we should aspire to create a machine learning driven manycore design framework. By doing so, we can overcome the design challenges in manycore systems and begin to create holistically optimized manycore systems that are able to fully-adapt to changing conditions. Such a framework can bring engineering costs down, commoditizing and democratizing the optimized computing platforms for the next generation of machine learning algorithms. Using this framework, we can stimulate the relationship between manycore systems and machine learning, spurring improvement in both computing infrastructures and algorithms.

Ryan Gary Kim is a Postdoctoral Research Associate in the ECE Dept. at Carnegie Mellon University. He got his PhD from Washington State University (2016). His current research interests are on the energy efficiency and scalability of manycore systems, machine learning for electronic design automation and computer architecture, and fully-adaptive system design. He is a member of the IEEE.

Janardhan Rao Doppa is an Assistant Professor in the School of EECS at Washington State University. He got his PhD from Oregon State University (2014) and his Masters from IIT Kanpur (2006). His research interests are machine learning, and data-driven science and engineering with a special focus on electronic design automation and computer architecture. He is member of ACM, IEEE, and AAAI.

Partha Pratim Pande is a Professor and holder of the Boeing Centennial Chair in computer engineering at the school of Electrical Engineering and Computer Science, Washington State University, Pullman, USA. His current research interests are novel interconnect architectures for manycore chips, on-chip wireless communication networks, and hardware accelerators for biocomputing. He received his PhD from the University of British Columbia in 2005. He is a senior member of IEEE.

Diana Marculescu is a Professor in the ECE Dept. at Carnegie Mellon Univ. She has won several best paper awards in top conferences and journals. Her research interests include energy-, reliability-, and variability-aware computing and CAD for non-silicon applications. She is an IEEE fellow.

Radu Marculescu is a Professor in the Dept. of Electrical and Computer Engineering at Carnegie Mellon University. He received his Ph.D. in Electrical Engineering from the University of Southern California (1998). He has received multiple best paper awards, NSF Career Award, Outstanding Research Award from the College of Engineering (2013). He has been involved in organizing several international symposia, conferences, workshops, and tutorials, as well as guest editor of special issues in archival journals and magazines. His research focuses on design methodologies for embedded and cyber-physical systems, biological systems, and social networks. Radu Marculescu is an IEEE Fellow.

Email Contacts

Ryan Gary Kim: rgkim@cmu.edu
Janardhan Rao Doppa: jana@eecs.wsu.edu
Partha Pratim Pande: pande@eecs.wsu.edu
Diana Marculescu: dianam@cmu.edu
Radu Marculescu: radum@cmu.edu